%% file: main_arxiv.tex
\documentclass{article}

\usepackage{microtype}
\usepackage{graphicx}
\usepackage{subfigure}
\usepackage{booktabs} 

\usepackage{hyperref}

\usepackage[accepted]{icml-preprint}

\usepackage{amsmath}
\usepackage{amssymb}
\usepackage{mathtools}
\usepackage{amsthm}
\usepackage{multirow}
\usepackage{hyperref}

\usepackage{makecell}

\usepackage[capitalize,noabbrev]{cleveref}

\theoremstyle{plain}

\theoremstyle{definition}

\theoremstyle{remark}

\DeclareMathOperator*{\argmax}{\mathrm{argmax}}

\newcommand{\chat}{GPT-3.5-Turbo}
\newcommand{\gptf}{GPT-4}

\newcommand{\si}[2]{#1^{(#2)}}

\usepackage[textsize=tiny]{todonotes}

\icmltitlerunning{Learning Planning-based Reasoning via Trajectories Collection and Process Reward Synthesizing}

\begin{document}

\twocolumn[
\icmltitle{Learning Planning-based Reasoning by \\Trajectories Collection and Process Reward Synthesizing}



\icmlsetsymbol{equal}{*}

\begin{icmlauthorlist}
\icmlauthor{Fangkai Jiao}{ntu,astar}
\icmlauthor{Chengwei Qin}{ntu}
\icmlauthor{Zhengyuan Liu}{astar}
\icmlauthor{Nancy F. Chen*}{astar}
\icmlauthor{Shafiq Joty*}{ntu,salesforce}
\end{icmlauthorlist}

\icmlaffiliation{ntu}{Nanyang Technological University, Singapore}
\icmlaffiliation{astar}{Institute for Infocomm Research, A*STAR, Singapore}
\icmlaffiliation{salesforce}{Salesforce Research, USA}

\icmlcorrespondingauthor{Nancy F. Chen*}{nyfchen@i2r.a-star.edu.sg}
\icmlcorrespondingauthor{Shafiq Joty*}{sjoty@salesforce.com}

\icmlkeywords{Large Language Models, Logical Reasoning, Process Supervision}

\vskip 0.3in
]



\printAffiliationsAndNotice{}  

\input{latex/abstract}

\input{latex/intro}

\input{latex/related_work}

\input{latex/method}

\input{latex/experiment}

\input{latex/results}

\input{latex/conclusion}


\nocite{langley00}

\bibliography{custom}
\bibliographystyle{icml/icml2024}

\newpage
\appendix
\onecolumn
\input{latex/appendix}

\end{document}

%% file: latex/abstract.tex
\begin{abstract}
Large Language Models (LLMs) have demonstrated significant potential in handling complex reasoning tasks through step-by-step rationale generation.
However, recent studies have raised concerns regarding the hallucination  and flaws in their reasoning process.
Substantial efforts are being made to improve the reliability and faithfulness of the generated rationales. Some approaches model reasoning as planning, while others focus on annotating for process supervision.
Nevertheless, the planning-based search process often results in high latency due to the frequent assessment of intermediate reasoning states and the extensive exploration space. Additionally, supervising the reasoning process with human annotation is costly and challenging to scale for LLM training.
To address these issues, in this paper, 
we propose a framework to learn planning-based reasoning through Direct Preference Optimization (DPO) on collected trajectories, which are ranked according to our synthesized process rewards.
Our results on challenging logical reasoning benchmarks demonstrate the effectiveness of our learning framework, showing that our 7B model can surpass the strong counterparts like \chat. 
\footnote{Code and trajectory data are released at \href{https://github.com/SparkJiao/dpo-trajectory-reasoning}{SparkJiao/dpo-trajectory-reasoning}.}
\end{abstract}

%% file: latex/intro.tex
\section{Introduction}
\label{sec:intro}
\input{images/example}

Natural language reasoning has been a fundamental element in the advancement of artificial intelligence (AI), with its significant impact on a variety of applications including planning and decision making~\citep{reasoning-llm-survey}. The goal of building AI systems capable of replicating human-like reasoning remains a primary focus within the research community.
Recent advancements in Large Language Models (LLMs) have showcased their ability to perform complex reasoning tasks, creating sequences of reasoning steps akin to human thought processes~\citep{cot,least-to-most}. Despite these advancements, it is also  concerning that LLMs are susceptible to generating misleading rationales \citep{gpt4-sparks,cot-faithful-anthrophic}. Such inaccuracies are particularly pronounced in complex reasoning scenarios \citep{alert,reasoning-llm-survey,logic-llm,seaeval}, underscoring a significant challenge.

Tremendous efforts have been dedicated to improve the reliability and faithfulness of generated rationales, including knowledge distillation~\citep{kd2015hinton,wizardllm,wizardmath,mammoth} and self-correction~\citep{shinn2023reflexion}.
Yet, these approaches predominantly rely on LLMs for identifying errors or providing superior reasoning processes, which could be limited by their capacity. 
An alternative is to consider \emph{human} process supervision~\citep{prm-and-orm}. For instance, \citet{lets-verify} propose to train a process reward model (PRM) using step-level feedbacks on model-generated solutions, which are annotated by human experts. This enables LLMs to refine their rationales based on the PRM's feedback. While human process supervision has proven effective, it often incurs higher costs compared to mere final outcome annotation as well as the automatic process annotation from a teacher LLM.
In addition to the attempts on process supervision, some research efforts have explored search-augmented reasoning for better reasoning trace by assessing the quality of future states.
\citet{rap} introduce a general framework of reasoning-as-planning (RAP), where the reasoning process is defined as a Markov Decision Process (MDP). 
{Each step in the MDP comprises a \textit{state}-\textit{action} pair, whose particular implementation can vary with different application scenarios.}
Figure~\ref{fig:example} illustrates this process in the context of logical reasoning using the ReAct~\citep{react} format. At each step the agent can  \emph{Think} (optionally) and \emph{Act}, which involves selecting a group of facts and rules to deduce a new conclusion.\footnote{The notion of \emph{action} subsumes both thinking and acting.} It can optionally make an \textit{Observation} to get an ``updated view'' of the \textit{state}. 
During inference, each state-action pair is assigned a reward, either by an LLM or  external verifier. The planning process is then steered by Monte Carlo Tree Search (MCTS) \citep{mcts} to maximize the {expected total cumulative} reward (or utility) obtained along the chosen path while effectively narrowing the search space {(Figure~\ref{fig:micro-frame}(a))}.

Existing RAP frameworks often assume LLMs as the world model being able to assess the quality of each reasoning step.
As a result, the online planning may introduce huge latency and cost due to frequent assessments of intermediate states and the large search space.
Nevertheless, we find that the core idea behind planning-based reasoning is to employ online simulation by taking few forward steps to find the optimal path, {and the evaluation becomes more accurate when it has access to real outcome feedback.}

\input{images/micro_framework}
In this paper, we explore \emph{offline simulation} to synthesize process supervision. We introduce ground-truth outcome supervision that we back-propagate to intermediate states instead of relying on LLMs  for process assessment.
We develop a simple and effective strategy based on partial trajectory exploration.
We first collect some solutions from LLMs as the seed trajectories, and then sample several intermediate reasoning states from them as the non-leaf nodes in planning. After that, the LLMs are asked to retry to complete each of them multiple times by taking the intermediate states as new starting points of reasoning.
We take the number of completions that have reached the correct outcome as the estimate of expected returns for training PRM. 
Finally, we optimize the LLMs to learn a better policy for generating reliable rationales through Direct Preference Optimization \cite{dpo}, where the contrastive trajectory pairs are annotated by the PRM.
A general comparison between our method and search-based approaches is shown in Figure~\ref{fig:micro-frame}.
In a nutshell, our contribution can be summarized as follows:
\begin{itemize}
    \item We propose a novel framework for synthesizing process rewards from outcome annotations, which incorporates offline simulation and  trajectory collection to induce planning-based reasoning.
    \item We rigorously evaluate our methodology on two challenging reasoning tasks: logical reasoning and mathematical reasoning. The observed significant improvements over robust baseline models underscore the efficacy of our proposed approach.
    \item Through detailed analysis, we demonstrate that our method not only improves the quality and conciseness of generated rationales but also reduces the reliance on human annotations.
\end{itemize}

%% file: images/example.tex
\begin{figure}[t]
    \centering
    \includegraphics[width=0.95\linewidth]{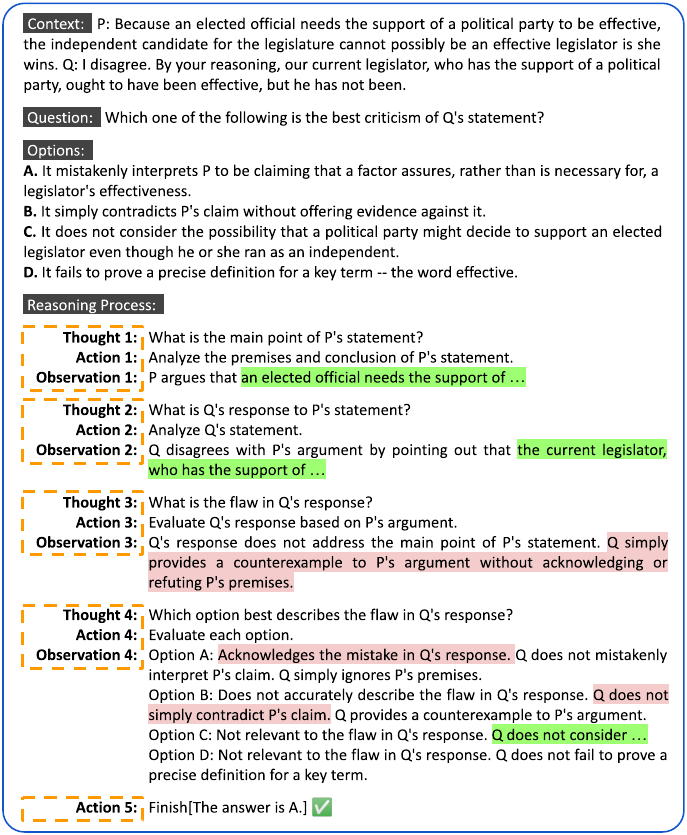}
    \vspace{-0.2cm}
    \caption{A solution generated by our fine-tuned model based on Llama-2-7B-chat~\citep{llama2} for a logical reasoning problem in LogiQA-v2 dataset~\citep{logiqa2}. It follow the ReAct~\citep{react} format, where {each step is marked with a dotted rectangle.} The content highlighted in green summarizes some opinions in the context, and is omitted. The central reasoning steps pivotal to arriving at the solution are emphasized in pink. The complete reasoning process can be found in Figure~\ref{fig:cmp-prompt}.}
    \label{fig:example}
    \vspace{-0.6cm}
\end{figure}

%% file: images/micro_framework.tex
\begin{figure}[t]
    \centering
    \includegraphics[width=0.95\linewidth]{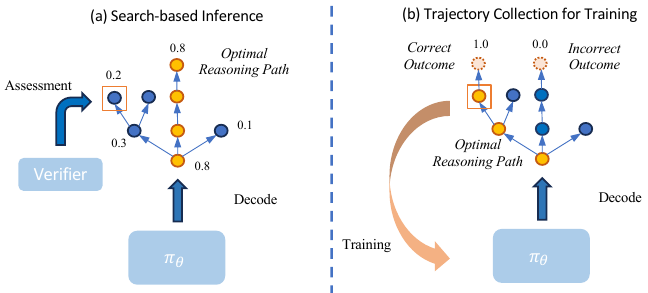}
    \vspace{-0.1cm}
    \caption{The overall comparison between search-based inference (a) and our trajectory collection-based offline training (b). In search-based inference, an LLM or an external verifier assesses each intermediate state and assigns a scalar value as feedback.
    The goal of inference is find an optimal reasoning path with maximum expected utility. In our method, the policy model will first explore multiple reasoning paths, with the process rewards calibrated by outcome supervision. And we then optimize it {using DPO \citep{dpo}} to maximize the probability of the paths with higher cumulative reward.  
    }
    \label{fig:micro-frame}
    \vspace{-0.6cm}
\end{figure}

%% file: latex/related_work.tex
\section{Related Work}
\label{sec:related-work}

\subsection{LLMs for Reasoning}


Compared with predicting only the final answer, chain-of-thought (CoT)~\citep{cot} serves as a more suitable way for LLMs considering the rationale will derive more useful information to avoid potential flaws.
Following this, many prompting techniques are proposed to enrich the generated rationales~\citep{least-to-most, rap}. 
Another group of work focuses on search-augmented reasoning, where the decoding process is guided by heuristic search algorithms, like MCTS~\citep{mcts}.
Basically, each reasoning state is treated as a node in a tree or graph, and assigned with a value demonstrating the confidence or expected rewards when reaching it. And LLMs themselves often serve as the evaluator to give feedback to intermediate states~\citep{tree-of-thought,rap}. 

\subsection{Improving LLMs via Sparse Feedback}

Since the success of reinforcement learning from human feedback (RLHF)~\citep{rlhf1,instructgpt}, employing RL algorithms, like PPO~\citep{ppo-openai}, to optimize LLMs from sparse feedback is becoming more important. However, PPO training often demonstrates unstable process and high resource cost. Some alternative variants are then proposed, like rejection sampling~\citep{constitutional-ai,llama2} and direct preference modeling (DPO)~\citep{dpo}.
Towards the different types of feedback, \citet{verify-step-by-step} and \citet{prm-and-orm} propose process supervision to assess the intermediate reasoning steps. Nevertheless, collecting step-wise feedback from human experts is often time-consuming and expensive. 
In this paper, we propose a simple heuristic approach to estimate the process rewards of intermediate states.

Our work is concurrent to MATH-Shepherd~\citep{math-shepherd}. We share similar methodology for process rewards estimation, but we have focused on different reasoning tasks, optimization approaches, and evaluations. More details are discussed in Appendix~\ref{app:shepherd}.


%% file: latex/method.tex
\input{images/framework}

\section{Method}
\label{sec:method}




 
\subsection{Formal Definition of Natural Language Reasoning}
\label{sec:definition}

Following~\citet{rap}, we define the natural language reasoning task as a MDP with an action-state trajectory: $\tau=\langle\,s_0, a_0,\cdots,s_t,a_t,\cdots,s_T,a_T\,\rangle$, where $a_t$ is the action  taken at timestep $t$ and $s_{t+1}$ is the state that the agent observes after that. 
In the context of LLMs, we simplify the setting by considering that both the action and state are sampled from the policy model $\pi_\theta$ (an LLM), such that:
\begin{equation}
\small{
    \begin{cases}
        &a_t\sim \pi_\theta(a|c_t), \\
        &s_{t+1}\sim \pi_\theta(s|a_t,c_t), \\
    \end{cases}
}
\end{equation}
where $\theta$ is the parameter of the policy model, $c_t=(s_0,a_0,\cdots,s_t)$ is the history trace. Besides, a reward model $r_t=r(a_t,s_t)\in\mathbb{R}$ is employed to assess the feasibility and desirability of each state-action pair.
In this paper, we focus on the tasks with annotated final labels, where the agent will receive a positive reward when it finally reaches a correct answer:
\begin{equation}
\small{
\label{eqn:full-reward}
    r_f(\tau, y)=\begin{cases}
        1, & \text{ if } \tau \to y   \\
        0, & \text{ else }
\end{cases}
}
\end{equation}
where $y$ is the ground-truth answer of a given query, and $\tau \to y$ means the trajectory entails the prediction $y$.
Our aim is to optimize the policy for making decisions to maximize the expected rewards, which can be formulated as:
\begin{equation}
    \argmax_{\theta} \mathbb{E}_{x,y\sim \mathcal{D},\tau'\sim \pi_\theta (\tau|x)} r_f(\tau',y),
\end{equation}
where $\pi_\theta$ is the policy model parameterized by $\theta$, $\mathcal{D}=\{\si{x}{i},\si{y}{i}\}$ is the dataset where the policy model is optimized on, $x$ is the concatenation of prompt, context, and question, and $\tau'$ is the generated reasoning process as action-state sequence.

\subsection{Estimate Process Rewards via Offline Simulation}
\label{sec:reward-collect}

One of the main issues with LLMs is that they  tend to  hallucinate \citep{survey-hallucinate}. A common illusion  with multi-step reasoning is that the derived conclusion may be correct but the LLMs might reach there through unreasonable deduction processes. To address this, we aim at introducing process supervision \cite{lets-verify}, which, however, is hard to obtain in most reasoning cases.
We propose a simulation based method to estimate the expected value by starting from an intermediate point in a trajectory and exploring the received rewards after reaching the terminal states.
{The idea is based on a common observation that if an intermediate reasoning state can reach the correct answer more frequently, it has higher probability to demonstrate some important facts or evidences towards the conclusion.}
Specifically, given an input $x$ and an intermediate reasoning step $t$, we randomly sample $K$ trajectories starting from either action $a_t$ or state $s_t$. Taking $a_t$ as example, the estimated expected value for it is formulated as:
\vspace{-0.3cm}
\begin{equation}
\small{
    \begin{aligned}
        &r_e(\tau_{t,a}, y) = \sum_{k}^K r_f(\tau^{k|\tau_{t,a}},y), \\
        =&\sum_{k}^K r_f(\langle\,\underbrace{s_0,a_0,\cdots,a_t}_{\text{{The prefix of $\tau$}}},\underbrace{s_{k,t},\cdots,s_{k,T_k}}_{\text{{The sampled completion.}}}\,\rangle,y),
    \end{aligned}
}
\vspace{-0.2cm}
\end{equation}

where $\tau^{k|\tau_{t,a}}$ is the $k$-th completed trajectory starting from $a_t$, and $T_k$ is the number of steps in the trajectory.
Note that we can estimate the expected value for both action or state, since they are all managed by the policy model. For simplicity, we will discuss the method based on action.

\subsection{Synthesized Process Reward Model}
\label{sec:prm}

After collecting enough trajectories as well as the estimated expected values of intermediate steps, we can train a PRM to assign a reward to each intermediate state/action, following \citet{verify-step-by-step}. The motivation behind training a process reward model instead of using the collected values as the rewards includes: (1) If we assess each intermediate step to estimate the value of the complete trajectory by only heuristic simulation, similar to the weakness of MCTS, the time consumption and cost will be severe. (2) The simulation based estimation will also introduce noise, since the completion quality highly depends on the fundamental capability of the initial policy model. As a result, employing an extra reward model to approximate the expected values can be more robust and efficient than heuristic algorithms.

Specifically, following the method in  Section~\ref{sec:reward-collect} we obtain a reward modeling dataset:
\begin{equation}
\small{
    \mathcal{D}^R=\{ \si{x}{i},\{ \si{\tau}{i}_{j,a},\si{r}{i}_{j} \} \},
}
\end{equation}
where $\si{r}{i}_{j}=r_e(\si{\tau}{i}_{j,a}, \si{y}{i})$, and $j$ is the step.
We then formulate the reward modeling process as a classification problem with $K$ classes, and train the process reward model $f_{\mathrm{prm}}:\mathcal{X}\times\mathcal{T}\rightarrow\mathbb{R}^{K}$ by minimizing the following Cross-Entropy loss:
\begin{equation}
\small{
    \begin{cases}
        &\mathcal{L}_{\mathrm{step}}=-\log p_r,    \\
        &p=f_\mathrm{prm}(x,\tau),
    \end{cases}
}
\end{equation}
where $\tau$ is an (incomplete) trajectory and $r$ is the corresponding estimated real reward value.

\subsection{Reward Annotation and Preference Dataset Construction}
\label{sec:dpo-dataset}

After obtaining the process rewards, we can then assess a complete trajectory by accumulating them along steps. Specifically, given a complete trajectory $\tau=\langle\,s_0,a_0,s_1,a_1,\cdots,s_T,a_T\,\rangle$, the trajectory level reward is defined as the accumulated production of the process rewards assigned at each intermediate step:
\begin{equation}
\small{
    \begin{cases}
        &r_p(\tau)=\prod_{t}^{T}\prod_{*}^{\{a,s\}} \sum_{i\geq C}^K f_\mathrm{prm}(\tau_{t,*})_{i}, \\
        &\tau_{t,a}=\langle\,s_0,a_0,\cdots,s_t,a_t\,\rangle, \\
        &\tau_{t,s}=\langle\,s_0,a_0,\cdots,s_t,\,\rangle,
    \end{cases}
    \label{eqn:traj-reward}
}
\end{equation}
where $*$ indicates either $a$ or $s$. $C$ is a hyper-parameter controlling the minimum amount of successful simulations so that we have enough confidence to claim the state can lead to a correct reasoning process. This is to avoid that the potential hallucinated rationales generated by the original LLMs can affect the estimation of process rewards.

Once we have the clear definition of the trajectory level reward based on the PRM, the policy model can be optimized via reinforcement learning. Considering the instability of PPO~\citep{ppo-openai} training, we choose the algorithm of Direct Preference Optimization (DPO) instead.

\subsection{Direct Preference Optimization}
\label{sec:dpo}


In this section, we will first introduce the vanilla DPO approach with only outcome supervision, which also servers as an strong baseline method. Specifically, given an original data sample $(\si{x}{i},\si{y}{i})$, and a group of trajectories $\si{\mathcal{T}}{i}=\{\si{\tau}{i}_{0},\si{\tau}{i}_{1},\cdots,\si{\tau}{i}_{n}\}$ sampled from the policy model taking $\si{x}{i}$ as input, we can simply construct a preference dataset:
\begin{equation}
\small{
    \mathcal{D}_\mathrm{o}=\{ \si{x}{i},\si{\tau}{i}_{w}, \si{\tau}{i}_{l} \},
    \label{eqn:dpo}
}
\end{equation}
where $\si{\tau}{i}_w\in\si{\mathcal{T}}{i}$ is a trajectory successfully reaching the correct answer $\si{y}{i}$, and $\si{\tau}{i}_{l}\in\si{\mathcal{T}}{i}$ is another trajectory with incorrect prediction. After that, we can optimize the policy model $\pi_\theta$ on the dataset $\mathcal{D}_\mathrm{o}$ by minimizing the following loss:
\begin{equation}
\small{
    \begin{aligned}
        &\mathcal{L}_{\mathrm{DPO}}(\pi_\theta; \pi_{ref}; \mathcal{D}_\mathrm{o}) \\
        =&  -\mathbb{E}_{(x,\tau_w,\tau_l)\sim \mathcal{D}_\mathrm{o}} \bigg[ \log \sigma \bigg( \beta \log \frac{\pi_\theta(\tau_w | x)}{\pi_{\mathrm{ref}}(\tau_w | x)} \\
& - \beta \log \frac{\pi_\theta(\tau_l | x)}{\pi_{\mathrm{ref}}(\tau_l | x)} \bigg) \bigg],
    \label{eqn:prm-filter}
    \end{aligned}
}
\end{equation}
where $\pi_\mathrm{ref}$ is the reference model initialized from the original policy model before DPO training, $\beta$ is the hyper-parameter controlling the divergence between the distribution from the policy model and the reference model, $\tau_w$ is the chosen solution, and $\tau_l$ is the rejected solution.

From the definition we can find that the vanilla DPO approach only considers the pairwise relationship based on final prediction, regardless of the reliability of intermediate reasoning process. Since we have already defined a trajectory-level reward in Equation~\ref{eqn:traj-reward} involving the process rewards, we can further consider the pair-wise relationship among those trajectories with correct predictions:
\begin{equation}
\small{
    \mathcal{D}_{\mathrm{p}}=\{ \si{x}{i}, \si{\tau}{i}_a, \si{\tau}{i}_b | r_p(\si{\tau}{i}_a) - r_p(\si{\tau}{i}_b) > \sigma, \},
}
\end{equation}
where
$\si{\tau}{i}_a$ and $\si{\tau}{i}_b$ both induce the correct prediction $\si{y}{i}$, and $\sigma$ is hyper-parameter representing the confidence margin. $\tau_a$ is the chosen solution and $\tau_b$ is the rejected one. And the final objective can thus be written as $\mathcal{L}_{\mathrm{DPO}}(\pi_\theta; \pi_{ref}; \mathcal{D}_\mathrm{o}\cup\mathcal{D}_\mathrm{p})$.

%% file: images/framework.tex
\begin{figure*}
    \centering
    \includegraphics[width=0.95\textwidth]{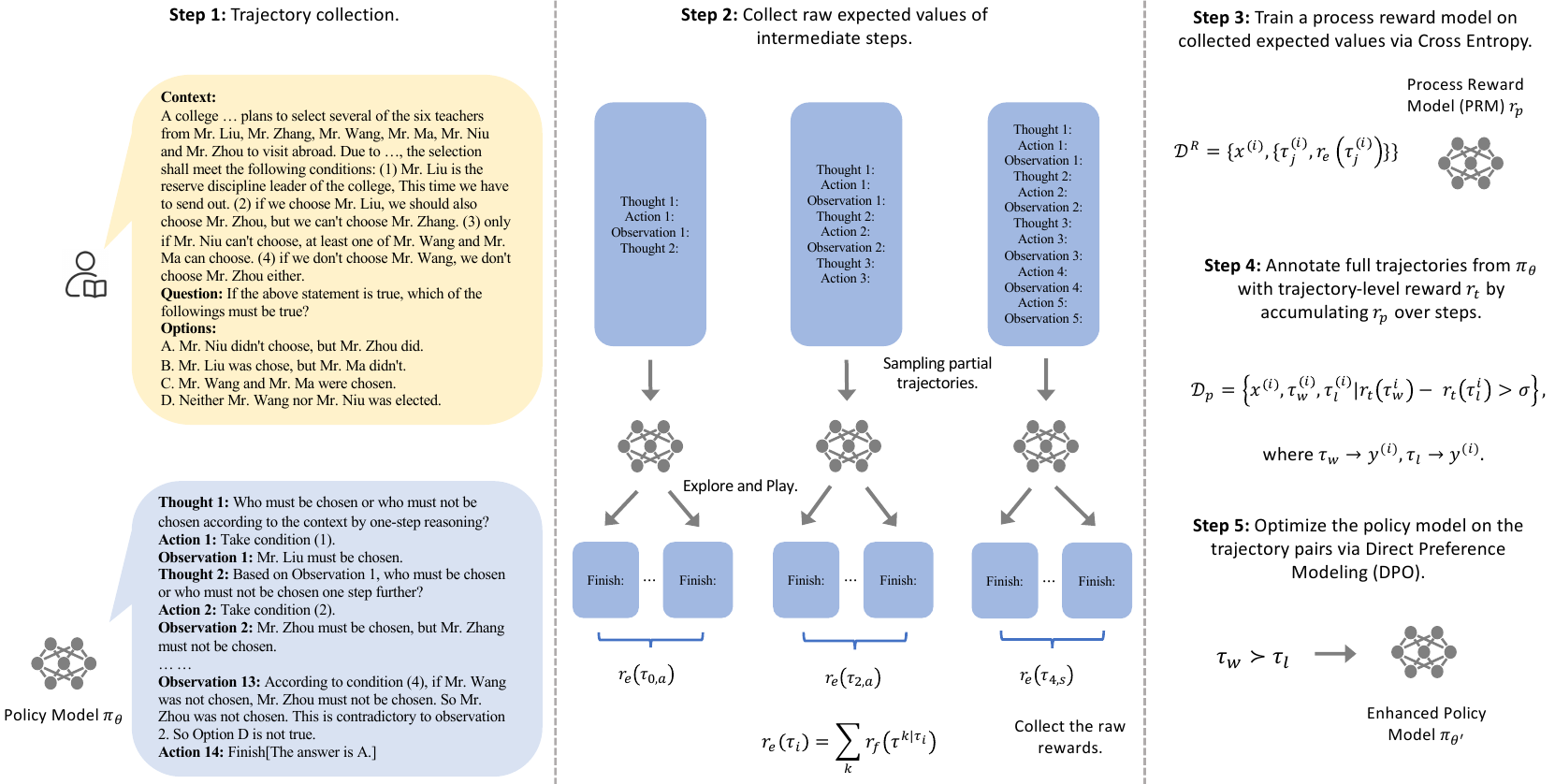}
    \caption{The overall framework of our approach. (1) Collect samples with full solution trajectories. (2) Sample intermediate reasoning states from the dataset, and ask the policy model to continuously explore based on the intermediate states. After the completed trajectory reaching the termination, we can collect the raw rewards according to the outcome supervision as the approximation of expected returns for the intermediate reasoning states. (3) A process reward model is learned from the raw rewards to alleviate the dataset noise and reduce simulation cost. (4) Collect more full trajectories and annotate them with the trained process reward  model. (5) Optimize the policy model on the pairwise trajectory dataset assessed by our synthesised process rewards.}
    \label{fig:framework}
    \vspace{-0.55cm}
\end{figure*}

%% file: latex/experiment.tex
\section{Experiments}

\subsection{Datasets}

In this paper, we mainly focus on logical reasoning and mathematical reasoning.
For logical reasoning, we choose ReClor~\citep{reclor} and LogiQA-v2~\citep{logiqa2} for evaluation, which are two challenging and widely used logical reasoning benchmarks. Both datasets are formulated as multiple choice question answering and the statistics of the two datasets are shown in Table~\ref{tab:dataset}.
For mathematical reasoning, we have employed the test sets of GSM8K~\citep{gsm8k} and MATH~\citep{hendrycksmath2021} for evaluation.

\subsection{Baselines}

For the baseline methods, we mainly choose the following types of approaches: (1) Foundational LLMs, including Llama2-70B-Chat~\citep{llama2}, Mixtral-MoE-8$\times$7B-Instruct~\citep{mixtral}, GPT-3.5-Turbo and GPT-4-Trubo\footnote{We use \textit{gpt-3.5-turbo-1106} and \textit{gpt-4-0125-preview}.}; (2) Supervised Fine-tuning (SFT), where the training solutions are sampled from larger teacher models; (3) DPO and IPO~\citep{ipo} methods with only outcome supervision; (4) Rejection-sampling based approaches, including Rejection-sampling based Fine-tuning (RFT)~\citep{RFT} and ReST-EM~\citep{rest-em}, and (5) reinforce learning (RL) algorithms for iterative training. The details of baselines can be found in Appendix~\ref{app:baseline}.

\subsection{Evaluation and Implementation}
\label{sec:eval}
The evaluation of open-ended generation is difficult. Considering that most of our models are fine-tuned on fixed format, we only take the solutions with satisfying the format requirements into consideration to calculate accuracy. The detailed rules can be found in Appendix~\ref{app:eval}. The implementation details can be found in Appendix~\ref{app:implementation}.

%% file: latex/results.tex
\input{tables/logical_reasoning_test}

\section{Results and Analysis}
\label{sec:exp}

\subsection{Overall Results on Logical Reasoning}

The results on logical reasoning benchmarks are shown in Table~\ref{tab:logic-reason}, from where we can conclude that
(1) DPO serves as a strong baseline, significantly boosting the performance of the SFT model and outperforming the other baselines. Notably, the DPO-fine-tuned model on LogiQA-v2 records an in-domain improvement of 7.0\%, and an 7.6\% improvement on the ReClor dataset. The one fine-tuned on ReClor also demonstrates 2.5\% in-domain and 3.0\% out-of-domain improvements, respectively. Besides, on LogiQA-v2, Llama2-7B-DPO can already surpass the other rejection sampling based baselines with large margins, like RFT and ReST-EM. This indicates DPO's efficacy in optimizing the policy model using outcome supervision alone.
(2) pDPO surpasses the vanilla DPO that relies solely on outcome supervision. For instance, by fine-tuning on LogiQA-v2, pDPO achieves absolute improvements of 2.4\% and 1.3\% on LogiQA-v2 and ReClor, respectively. Through training on ReClor, pDPO also achieves 2.2\% absolute in-domain improvements.
Besides, pDPO trained on LogiQA-v2 outperforms the strong foundation LLMs including Mixtral and GPT-3.5 Turbo, suggesting the superiority of our synthesized process supervision.
(4) The LogiQA-v2 dataset emerges as a more effective tool for learning explicit logical reasoning processes compared to ReClor. As shown in the Table, by fine-tuning on LogiQA-v2, the generalization performance of pDPO on ReClor dataset is even better than the in-domain fine-tuned models. After diving into the dataset details, we find that LogiQA-v2 comprises multiple complex logical reasoning abilities, like categorical reasoning and sufficient reasoning, while quite a few questions in ReClor require only one-step reasoning to justify the entailment of each option.

\subsection{Improvements by Iterative Training}
We also performed iterative training by taking Llama2-7B-pDPO trained on LogiQA-v2 as the new base model and fine-tuning it on the newly self-sampled solutions. In addition to DPO and pDPO, we have also explored the RL based approaches, including PPO~\citep{ppo-openai} and Group Relative Policy Optimization (GRPO)~\citep{deepseek-math}. For fair comparison with pDPO, PPO and GRPO also include both the process rewards from our PRM, and the outcome rewards derived from the ground-truth labels. The implementation details can be found in Appendix~\ref{app:baseline}.

From Table~\ref{tab:logic-reason}, we observe that all four approaches demonstrate consistent in-domain improvements. Notably, the pDPO approach, which utilizes synthesized process supervision, surpasses the conventional process PPO method. 
This improvement may be attributed to the slightly noisy nature of the synthesized process rewards, which complicates the task for the critic model within the PPO algorithm to accurately approximate the distribution and reduce the variance of the expected returns.
Conversely, GRPO achieves a significant performance edge over PPO by sampling multiple solutions for the same query and calculating advantages using the group-averaged rewards as baseline.
Furthermore, it is important to highlight that DPO-based methods significantly reduce training costs, completing the training process in under 16 hours on four NVIDIA H100 GPUs, whereas PPO and GRPO require over 40 hours on the same hardware.

\input{tables/math_reasoning}

\subsection{Results on Mathematical Reasoning}

In addition to logical reasoning, we also conducted experiments on mathematical reasoning to verify the effectiveness of our proposed approach, and the results are shown in Table~\ref{tab:math}. 
Specifically, we randomly sampled a subset of MetaMath~\citep{metamath} as the training set containing 25,000 questions for Gemma-2B training. 
From the table we can conclude that, on GSM8K, the synthesized process rewards also effectively enhance the mathematical reasoning capabilities. Moreover, by employing DPO and pDPO, our models with 2B parameters can outperform Gemma-7B-Instruct with significant improvements.

Despite our efforts, enhancing Gemma-2B-DPO's performance on the MATH dataset has proven challenging, possibly due to the base model's limited capability on MATH, which introduces noise when estimating expected returns during the simulation stage. Consequently, we expanded our experiments to include DeepSeekMath-7B-Instruct~\citep{deepseek-math}, which is pre-trained on large high-quality math-related corpus. 
We curated another subset from MetaMath for DeepSeekMath training, which contains 55,000 questions augmented from the MATH training dataset. 
As depicted in the table, the results reveal that pDPO also surpasses DPO in performance by employing better foundation model.

\input{images/data_ratio}

\subsection{Reliance on Annotations of Outcome Supervision}

Although our proposed reward synthesis approach have avoided the direct annotation of process supervision, the outcome supervision still plays an important role for back-propagating the confidence to intermediate reasoning steps. In order to study the effect of outcome supervision scale to final performance, we randomly construct the sub-datasets containing 40\%, 60\%, and 80\% questions in the original dataset to evaluate the fine-tuned performance. The results are plotted in Figure~\ref{fig:data-ratio}.

From the figure we can observe that 
(1) pDPO consistently outperform DPO across all dataset splits with different sizes by significant margins, demonstrating the effectiveness of the synthesized process supervision. 
(2) With only 40\% annotations containing 3,234 questions in total, process supervision can outperform the base SFT model with significant improvements, which also verifies the significance by providing sparse feedback for continuous improvements.
(3) Besides, we find that pDPO with only 40\% outcome annotations can achieve comparable performance on the test set with DPO, i.e., 53.5 v.s. 53.9. Considering that we have only used \textbf{10\%} outcome annotations for training the process reward model, the results can definitely emphasize the data efficiency of our approach.







\input{images/reward_analysis}

\input{images/win_rate}

\subsection{Auto-evaluation of Rationale Quality by GPT-4}

The most important concern is whether the synthesised process reward can contribute to reasonable and reliable rationale generation. In order to evaluate this,
we propose to use GPT-4 for automatic evaluation. Specifically, following \citet{lima} and \citet{chatbot-arena}, we first formalize three dimensions to assess the rationales: \textit{Reasonable}, \textit{Concise}, and \textit{Logically Consistent}.
We then give \gptf~two reasoning process and ask it to judge which one is better or it is a tie for each aspect. The critique details and prompt are shown in Figure~\ref{fig:cmp-prompt}.
In order to avoid the bias caused by prediction errors of the two models, we first find a subset of questions where both the solutions given by the two models lead to the correct answer. After that, we randomly sampled 261 questions from the subset for evaluation. The results are shown in Figure~\ref{fig:win-rate}. From all the three aspects, pDPO performs much better than vanilla DPO without process reward. Around 67.8\% solutions of pDPO are deemed to have higher overall quality. Besides, for the most important view, among 52.5\% questions, pDPO can generate more reasonable rationales. 
We can also find that nearly 60\% responses by pDPO are more compact, suggesting that the process supervision can help make the rationale more brief but accurate.

\subsection{Analysis of Predicted Rewards}

In this section, we have visualized the predicted step-wise rewards on the training set of LogiQA-v2, where the solutions are sampled from the SFT model. In Figure~\ref{fig:rewards}, we have visualized three different kinds of rewards: (1) the averaged step-wise rewards before the softmax operation, i.e., the probability (left); (2) the accumulated rewards by production (medium); and (3) the averaged logits of each step from the last layer of the reward model (right).
When diving into the logits without normalization, we can find that the rewards maintain relatively stable at around the first 15 steps, then decrease sharply. This may be caused by the imbalanced amount of solutions with different reasoning steps, which makes the reward model less confident on the longer steps. On the other hand, the accumulated probability based rewards keep decreasing with longer reasoning process, which can be useful to avoid redundant solutions by penalizing the extremely longer ones.

\subsection{Case Study}

In this section, we conduct a case study to intuitively demonstrate the augmentation bought by process-supervised DPO. As shown in Figure~\ref{fig:case}, the vanila DPO induced model shows two weaknesses: (1) the intermediate reasoning step is wrong, which is highlighted in red.
And (2) the solution is redundant, like \textit{Action 2} and \textit{Action 5} to \textit{Observation 8}.
On the contrary, process-supervised DPO not only well illustrates the flaw in Q's response (\textit{Observation 3}), but also eliminate the meaningless content, which introduce less noise to make correct prediction.

%% file: tables/logical_reasoning_test.tex
\begin{table}[t]
    \centering
    \renewcommand{\arraystretch}{1.2}
    \setlength{\tabcolsep}{2.0mm}{
    \scalebox{0.75}{
    \begin{tabular}{l|ccc}
    \toprule
                                & \makecell{Training \\ Set}    & LogiQA-v2  & ReClor \\\hline
    \chat                       & ---   & 45.4 & 53.7 \\
    \gptf-Turbo                 & ---   & \underline{70.0} & ---\\ \hline
    Llama2-70B-chat             & ---   & 43.8  & 60.4 \\
    Mixtral-8$\times$7B-Instruct
                                & ---   & 49.5   & 56.7 \\\hline
    {Llama2-7B-SFT}          & ReClor    & 44.5  & 48.8  \\ 
    {Llama2-7B-DPO}          & ReClor    & \textbf{47.5}  & 51.3  \\ 
    {Llama2-7B-pDPO}         & ReClor    & 47.4  & \textbf{53.5} \\ 
    \midrule
    Llama2-7B-SFT                   & LogiQA-v2 & 45.5  & 53.4 \\\hdashline
    Llama2-7B-RFT \\
    ~~~~Outcome              & LogiQA-v2 & 47.8 & 52.3 \\
    ~~~~Outcome \& PRM-top-1 & LogiQA-v2 & 48.0 & 54.2 \\ 
    ~~~~Outcome \& PRM-top-3 & LogiQA-v2 & 48.1 & 53.0 \\
    ~~~~Outcome \& PRM-top-5 & LogiQA-v2 & 47.9 & 52.6 \\\hdashline
    Llama2-7B-ReST-EM       & LogiQA-v2 & 49.4  & 51.5  \\
    ~~~~Iter-1       & LogiQA-v2 & 48.7  & 52.8 \\\hdashline
    Llama2-7B-IPO           & LogiQA-v2 & 44.5  & 54.1 \\
    Llama2-7B-DPO           & LogiQA-v2 & 53.1  & 60.4  \\
    Llama2-7B-pDPO          & LogiQA-v2 & \textbf{55.5} & \textbf{61.7} \\\hline  
    ~~~~Iter-1-DPO   & LogiQA-v2 & 56.7  & 61.0  \\
    ~~~~Iter-1-pDPO  & LogiQA-v2 & \textbf{57.3}  & \textbf{61.8}  \\
    ~~~~Iter-1-process PPO   & LogiQA-v2 & 56.2  & 61.2 \\
    ~~~~Iter-1-process GRPO  & LogiQA-v2 & \textbf{57.3}  & \textbf{61.7} \\
    \bottomrule
    \end{tabular}
    }}
    \vskip -0.05in
    \caption{Experimental results on the test set of the logical reasoning benchmarks.}
    \label{tab:logic-reason}
    \vspace{-0.6cm}
\end{table}

%% file: tables/math_reasoning.tex
\begin{table}
    \centering
    \renewcommand{\arraystretch}{1.0}
    \setlength{\tabcolsep}{2.0mm}{
    \scalebox{0.85}{
    \begin{tabular}{l|cc}
    \toprule
                        & GSM8K     & MATH  \\\hline
    Gemma-7B-Instruct   & 46.4      & 24.3  \\ \hdashline
    Gemma-2B-SFT        & 45.8      & 14.1  \\
    Gemma-2B-DPO        & 50.6      & \textbf{16.0} \\
    Gemma-2B-pDPO       & \textbf{52.8} & 15.7 \\ \hline
    DeepSeekMath-7B-Ins. & 82.3 & 45.1 \\
    DeepSeekMath-7B-Ins. + DPO  & \textbf{82.4}  & 46.3 \\
    DeepSeekMath-7B-Ins. + pDPO & 82.3  & \textbf{46.8} \\
    \bottomrule
    \end{tabular}
    }}
    \vskip -0.10in
    \caption{Experimental results on mathematical reasoning. \textit{Ins.} is the short for \textit{Instruct}, indicating we are using the instruction tuned version of DeepSeekMath. All experiments except the SFT one are repeated for \textbf{3 times} and the averaged results are reported.}
    \label{tab:math}
    \vspace{-0.6cm}
\end{table}

%% file: images/data_ratio.tex
\begin{figure}
    \centering
    \includegraphics[width=0.95\linewidth]{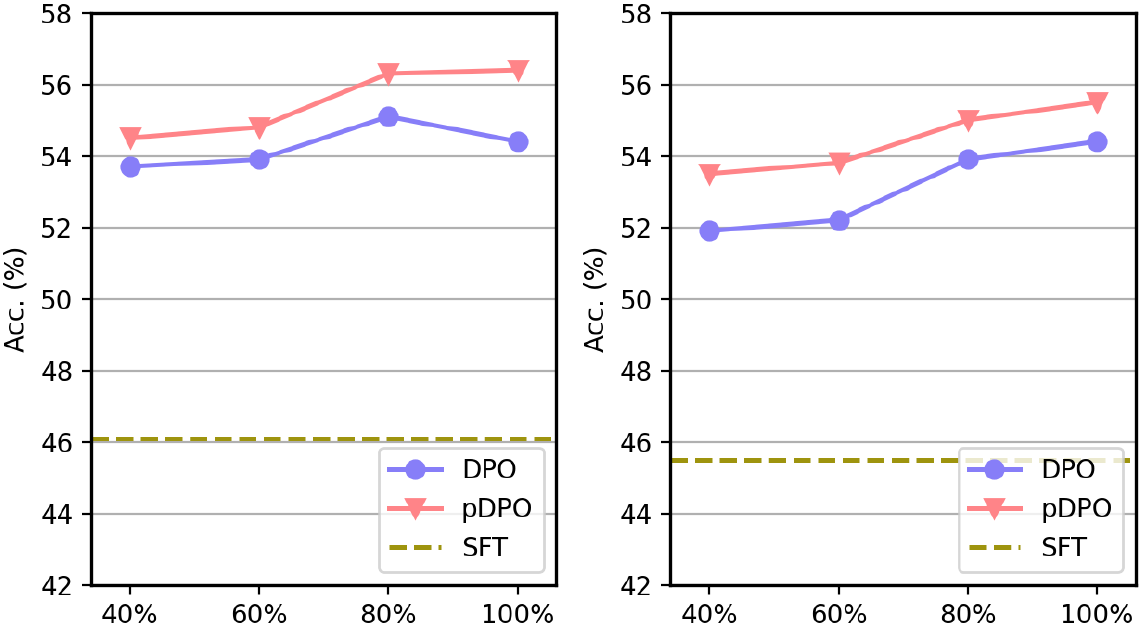}
    \vspace{-0.2cm}
    \caption{The accuracy of DPO, pDPO and SFT models on the validation set (left) and test set (right) of LogiQA-v2, respectively, taking different ratio of annotated questions.}
    \label{fig:data-ratio}
    \vspace{-0.7cm}
\end{figure}

%% file: images/reward_analysis.tex
\begin{figure*}
    \centering
    \includegraphics[width=0.95\textwidth]{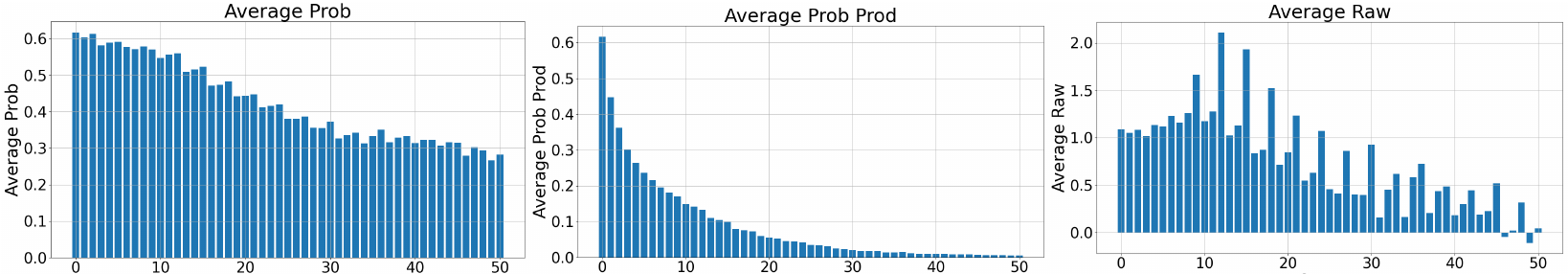}
    \vspace{-0.2cm}
    \caption{The averaged reward scores of intermediate reasoning steps predicted by our trained process-reward model on the training set of LogiQA-v2. The x-axis indicates the amount of reasoning steps and the y-axis describes the value of the averaged scores. For left to right, the three figures illustrate (1) predicted probability based reward of each reasoning step; (2) the accumulated probability based reward till specific reasoning step by production; and (3) the raw predicted reward values from the last layer of the reward model with different reasoning steps.}
    \label{fig:rewards}
    \vspace{-0.5cm}
\end{figure*}

%% file: images/win_rate.tex
\begin{figure}
    \centering
    \includegraphics[width=1.0\linewidth]{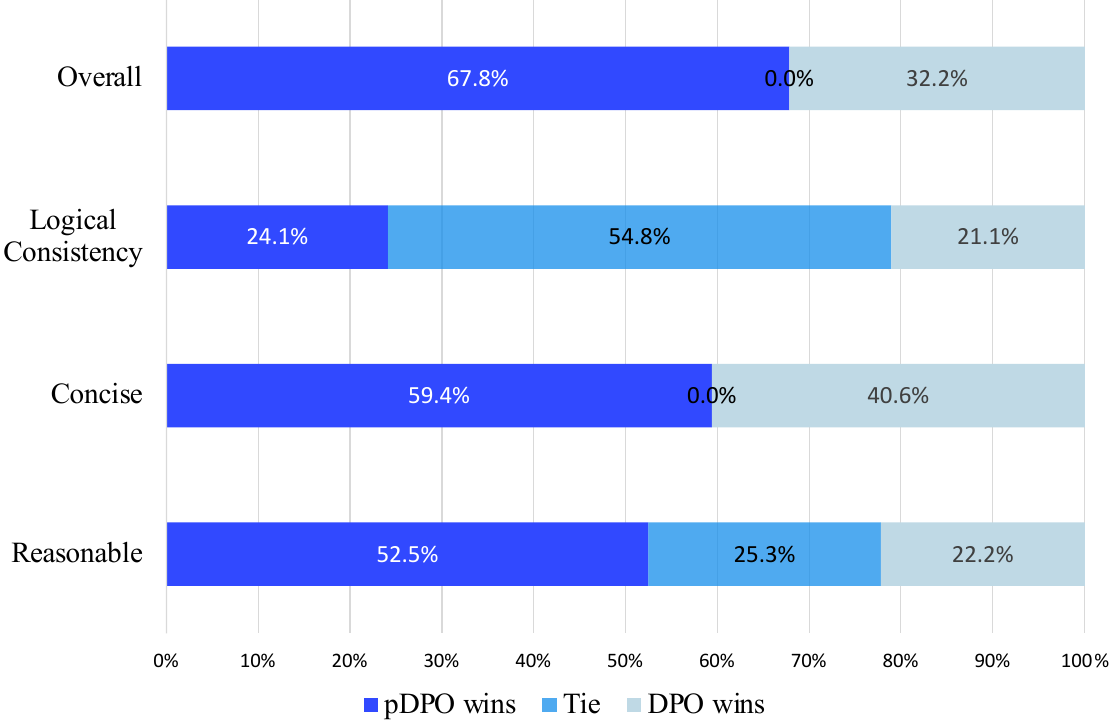}
    \vspace{-0.6cm}
    \caption{The wining rate between DPO and pDPO over different aspects of the auto-evaluation of \gptf.}
    \label{fig:win-rate}
    \vspace{-0.6cm}
\end{figure}

%% file: latex/conclusion.tex
\section{Conclusion}


In this paper, we propose a novel idea to transform reasoning-as-planning as a learning problem to avoid the latency induced by online search. Inspired by MCTS, we developed a offline simulation approach to estimate the expected value of intermediate reasoning steps. After that, we use the collected expected value dataset to fit a process reward model and annotate the full trajectories with sequence-level rewards. Finally, the policy model is optimized using direct preference optimization. The experimental results on logical and mathematical reasoning demonstrate the effectiveness of our proposed method. 
Towards the future work, we hope to explore the synthesised process reward estimated by weak-supervision from different aspects to further alleviate the reliance on human annotations and enable consistent self-improvement.

%% file: latex/appendix.tex
\section{Baseline}
\label{app:baseline}

\paragraph{Foundational LLMs} We have selected the strong LLMs without task-specific fine-tuning as baselines, including Llama-2-70B-chat~\citep{llama2}, Mixtral-MoE-8$\times$7B-Instruct~\citep{mixtral}, GPT-3.5-Turbo and GPT-4-Turbo~\citep{gpt4-tech-report}. 
\paragraph{Supervised Fine-tuning (SFT)} We first sample some responses from larger LLMs following the ReAct format for knowledge distillation since we cannot directly fine-tune them due to the resource limitation. After that, we can obtain the smaller LLMs with considerable reasoning capability through supervised fine-tuning (SFT). These models serve as baselines and the foundation models for DPO training. Specifically, we choose Llama-2-7B-chat and Gemma-2B-Instruct~\citep{gemma} for SFT.

\paragraph{Outcome-based Preference Optimization} We include the model with only outcome supervision as baseline to discuss the effectiveness of our synthesised process reward. For fair comparison, DPO implicitly model the outcome rewards following Equation~\ref{eqn:dpo}. We also involve IPO as baseline. The training dataset is $\mathcal{D}_o$ as mentioned in Section~\ref{sec:dpo}.

\paragraph{Rejection Sampling-based Approach} We also include the rejection sampling based approaches, i.e., Rejection sampling based Fine-tuning~\citep{RFT}, and ReST-EM~\citep{rest-em}. Both approaches use outcome annotations to filter the self-sampled solutions. The difference is that RFT uses the correct solutions to augment the original SFT dataset, while ReST-EM employs the sampled dataset to train the original model from scratch during each iteration.
Besides, for RFT, we includes two variants: (1) RFT-outcome uses only the outcome annotation to filter solutions; and (2) RFT-outcome \& PRM-top-k follows RFT-outcome and uses our trained PRM to rank the kept solutions. Only the top-k ranked solutions will be kept and augment the orinal training set.
For ReST-EM, we have conducted two iterations since there is already performance decreasing observed in the second round.

\paragraph{Reinforce Learning} In the experiments of iterative training, we include two reinforce learning algorithms, PPO~\citep{ppo-openai} and GRPO~\citep{deepseek-math} as the comparison of process-based DPO. Both algorithms employ two kinds of rewards, i.e., the outcome reward and the process rewards. For each solution (trajectory) sampled from the policy model, we assign it with $1$ if it can induce the correct answer, otherwise we assign it with $0$ as the outcome reward. Besides, for each reasoning step, the predicted logits by our trained PRM is treated as the process rewards. One difference should be noted is that, in pDPO training, we utilize the probability from the PRM as the process reward following \citet{lets-verify}, while for RL training, we use the logits without normalization from the last layer of PRM, to avoid extreme longer solutions introduced by accumulating the non-positive step rewards.

\section{Evaluation Details}
\label{app:eval}

In order to simplify the evaluation procedure, for the models without task-specific fine-tuning, we use 1-shot prompt of ReAct, which is the same as that we used for collecting data, to induce the models to generate reasonable solutions. For models after fine-tuning, we remove the 1-shot demonstration because we find it can lead to higher results. Due to limitation of budget, for GPT-4-Turbo, we only evaluate the first 250 questions in the test set of LogiQA-v2.

Besides, as mentioned in Section~\ref{sec:eval}, we have designed several rules to both filter the solutions unsatisfying the ReAct format and calculate the accuracy. Specifically, all of the following cases will be considered incorrect:
\begin{itemize}
    \item The final answer contains more than one prediction, e.g., \textit{Finish[The answer is A and B]}.
    \item The solution is truncated due to the length limit, but some option indices are gathered.
    \item The summary format is incorrect, e.g., \textit{Finish: the answer is A}.
\end{itemize}
For experiments with DeepSeekMath~\citep{deepseek-math} on mathematical reasoning, we only do basic cleaning like removing the redundant newline symbols, since it is already fine-tuned on the solutions with CoT format.

\input{tables/dataset}

\section{Implementation Details}
\label{app:implementation}

\subsection{Data Preparation}
Considering the limited computation resources,  we mainly conducted experiments on Llama2-7b-chat~\citep{llama2}, DeepSeekMath~\citep{deepseek-math}-7B-Instruct, and Gemma-2B-Instruct~\citep{gemma}. In order to collect solutions reaching correct answers more efficiently, we first fine-tune the original models on corresponding dataset using the generated responses from some teacher models (except DeepSeekMath since its solutions are already in CoT format). For LogiQA-v2, we sample solutions from Llama-2-70b-chat, while for ReClor, the solutions are sampled from GPT-3.5-Turbo to save time. For Gemma-2B, we sample solutions of MetaMath from Qwen-72B-chat~\citep{qwen}. 

All teacher models are prompted with exactly one example. The prompt used for LogiQA-v2 and ReClor is shown in Figure~\ref{fig:prompt}. And the one used for MetaMath follows RAP~\citep{rap}\footnote{\href{https://github.com/Ber666/RAP/blob/main/data/gsm8k/prompts/interactive_examples.json}{https://github.com/Ber666/RAP/data/gsm8k/prompts.}}. For all datasets, we sample 10 solutions regarding each question with temperature fixed as 0.7. Besides, for ReClor dataset, we remove all solutions with less than 8 reasoning steps because they omit the detailed reasoning process and can lead to inferior solutions for DPO based approach.

\subsubsection{Training Data Collection For PRM}

For LogiQA-v2, we randomly sampled 10\% questions from the training set for process rewards estimation and PRM training. For ReClor, the ratio is 20\%. For Gemma-2B training, we have used 25\% questions for PRM training, while for DeepSeekMath, we have used around 10\%.

\subsection{Hyper-Parameters}

For hyper-parameters, we use $\beta=0.1$ and $C=2$ on logical reasoning tasks, and $\beta=0.5$, $C=3$ on mathematical reasoning tasks.
Besides, $\sigma$ is set as 0.4 for ReClor dataset, 0.5 for LogiQA-v2, 0.5 for Gemma-2B, and 0.3 for DeepSeekMath.

\subsection{Training}

All experiments are conducted on NVIDIA A100 and H100. The evaluation of LLMs relies on vLLM~\cite{vllm} inference backend. 
For logical reasoning, after training, we evaluate all checkpoints on the development set of the target dataset using greedy decoding, and select the best one to report its performance on the test set.
For Gemma-2B, we select the model checkpoint based on the performance on GSM8K, and for DeepSeekMath, we report the performance of the best checkpoint on MATH.
All experiments, expept those using RL algorithms, are \textbf{repeated for 3 times} with different random seeds and the average results are reported to reduce the influence of randomness.
We run RL-based approaches for only once due to resource limitation.

\section{Compared with MATH-Shepherd}
\label{app:shepherd}
\input{latex/compare_math_shepherd}

\input{tables/sigma}
\section{Effect of Different Reward Margins}

In Equation~\ref{eqn:prm-filter}, we have involved a hyper-parameter $\sigma$ to control the confidence interval between different sample pairs both reaching the correct answer to construct the process-supervised preference dataset.
Naturally, there are several aspects of trade-off to considering the choices of $\sigma$. $\sigma$ with higher value can improve the ratio of true positive pairs in the constructed dataset. Yet, high confidence intervals will also reduce the number of training data and probability to include more hard negative samples.
For example, as shown in Table~\ref{tab:sigma}, $\sigma=0.7$ introduces only 10\% extra preference pairs and lead to less significant improvements compared with the case where $\sigma=0.5$.
On the other hand, lower value of $\sigma$ can include both more hard negative and false positive pairs. From the table we find that $\sigma=0.3$ has has introduced more than 25\% process-supervised pairs, but the performance is even worse than the vanilla DPO approach, where only outcome-based preferences pairs are employed.

\input{images/prompt}
\input{images/cmp_prompt}
\input{images/case}

%% file: tables/dataset.tex
\begin{table}
    \centering
    \renewcommand{\arraystretch}{1.2}
    \setlength{\tabcolsep}{1.0mm}{
    \scalebox{0.75}{
    \begin{tabular}{ccccc}
    \toprule
        Dataset     
                    & \makecell[c]{\# Question\\(Train)}
                    & \makecell[c]{Avg. of Correct\\Solutions\\Per. Question}
                    & \makecell[c]{\# Question\\(Val.)}
                    & \makecell[c]{\# Question\\(Test)}  \\ \hline
        LogiQA-v2   & 12,567        & 6.0       & 1,569     & 1,572   \\
        ReClor      & 4,638         & 5.0       & 500       & 1,000     \\
    \bottomrule
    \end{tabular}
    }}
    \caption{Statistics of our used datasets in this paper for construction preference pairs. The solutions shown in the table are sampled from the corresponding SFT model based on the questions in the training set.}
    \label{tab:dataset}
    \vspace{-0.2cm}
\end{table}

%% file: latex/compare_math_shepherd.tex
We work concurrently with Math-Shepherd~\citep{math-shepherd}, which also comprises similar offline simulation method to synthesize the process supervision. Differently, they mainly evaluate the approach on mathematical reasoning through verification, where the candidate solutions are ranked according to the rewards from the learned PRM, or employing it for PPO training, while we focus on logical reasoning and demonstrate the effectiveness of the synthesized process supervision via constructing the preference dataset under the guidance of the PRM. The dataset is further used for DPO training, which, though cannot really surpass GRPO, often demonstrates less resource requirements and more stable learning process.

%% file: tables/sigma.tex
\begin{table}
    \centering
    \renewcommand{\arraystretch}{1.2}
    \setlength{\tabcolsep}{1.5mm}{
    \scalebox{0.75}{
    \begin{tabular}{cccccc}
    \toprule
        $\sigma$    & No. of Pairs  & No. of P. Pairs  & Ratio of P. Pairs & Dev. & Test  \\ \hline
        1.0         & 133,458       & 0         & 0         & 54.4  & 54.4 \\ \midrule
        0.3         & 179,776       & 46,318    & 25.8\%    & 51.4  & 50.4   \\
        0.5         & 161,140       & 27,682    & 17.2\%    & \textbf{56.4}  & \textbf{55.5}  \\
        0.7         & 148,136       & 14,678    & 9.9\%     & 55.7  & 54.3  \\
    \bottomrule
    \end{tabular}
    }}
    \caption{Accuracy on LogiQA-v2 dataset with different $\sigma$. $\sigma=1.0$ refers to the vanilla DPO method. \textit{P. Pairs} refers to \textit{process-supervised sample pairs}.}
    \label{tab:sigma}
    \vspace{-0.5cm}
\end{table}

%% file: images/prompt.tex
\begin{figure*}
    \centering
    \includegraphics[width=1\textwidth]{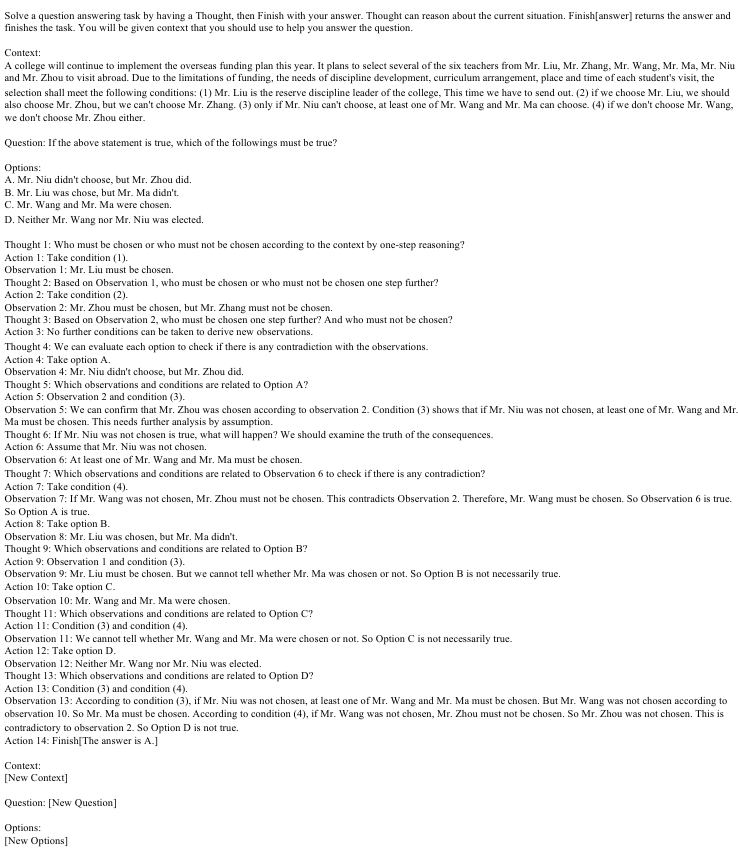}
    \caption{Prompt for sampling ReAct solutions. For zero-shot prompting, the exemplar is removed while the other content keep unchanged.}
    \label{fig:prompt}
\end{figure*}

%% file: images/cmp_prompt.tex
\begin{figure*}
    \centering
    \includegraphics[width=0.9\textwidth]{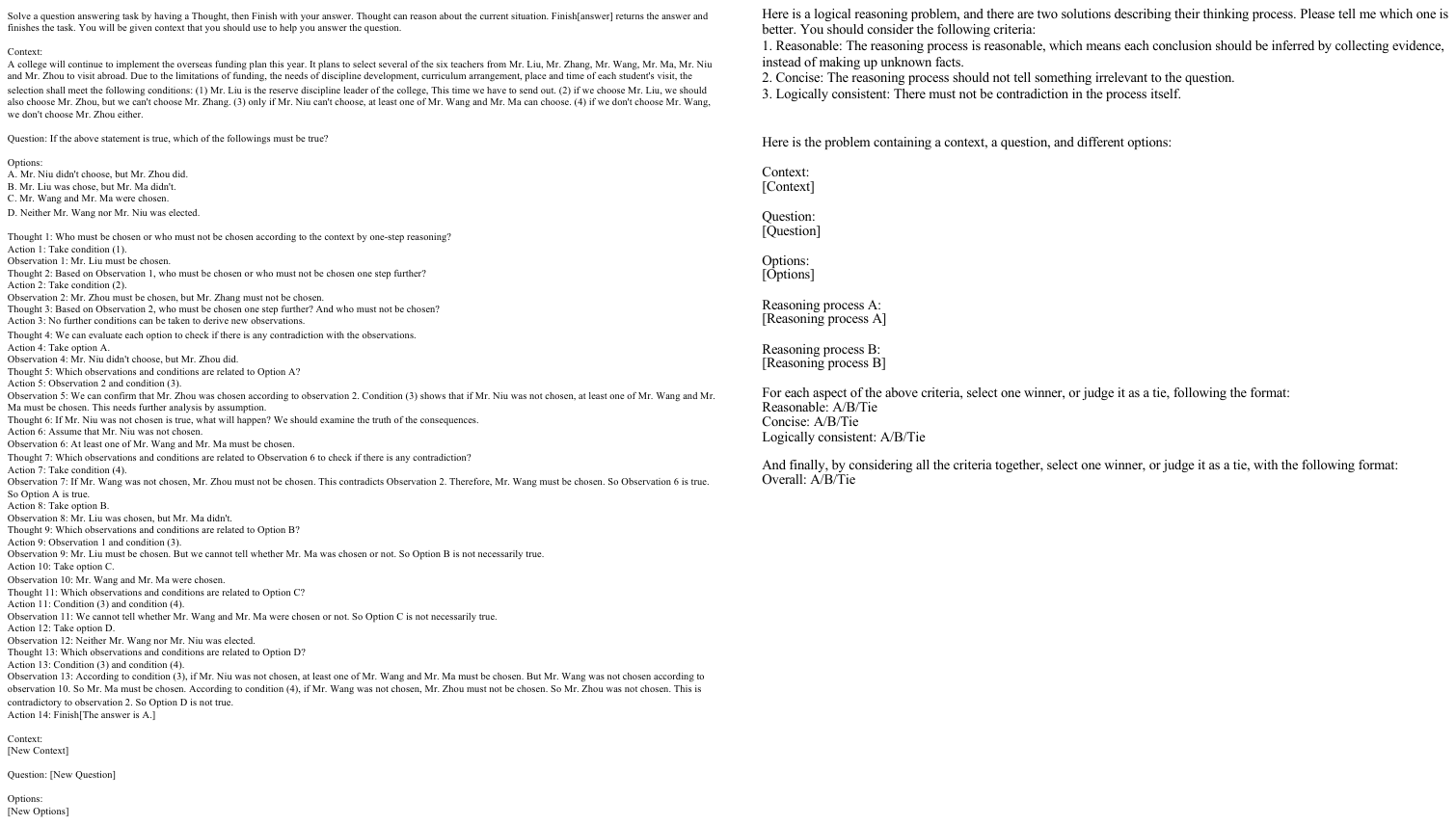}
    \caption{Prompt for compare two different responses, and assess the quality from different aspects.}
    \label{fig:cmp-prompt}
\end{figure*}

%% file: images/case.tex
\begin{figure*}
    \centering
    \includegraphics[width=0.9\textwidth]{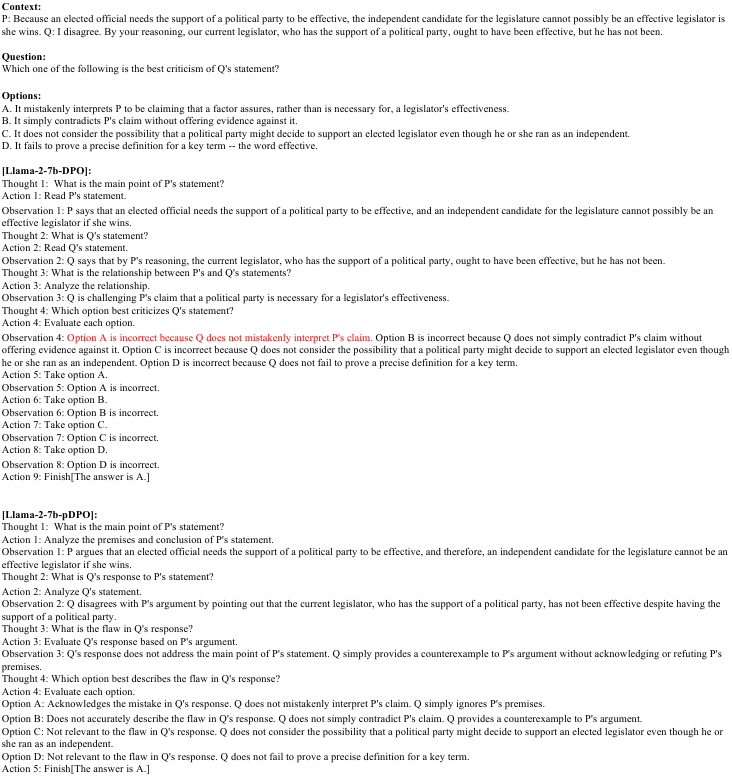}
    \caption{Case study taking the responses from Llama2-7b-DPO and Llama2-7b-pDPO that are fine-tuned on LogiQA-v2, respectively.}
    \label{fig:case}
\end{figure*}